\documentclass{article}
\usepackage{ijcai24}

\usepackage{times}
\usepackage{soul}
\usepackage{url}
\usepackage[hidelinks]{hyperref}
\usepackage[utf8]{inputenc}
\usepackage[small]{caption}
\usepackage{graphicx}
\usepackage{amsmath}
\usepackage{amsthm}
\usepackage{booktabs}
\usepackage{algorithm}
\usepackage{algorithmic}
\usepackage[switch]{lineno}

\usepackage{algorithm}
\usepackage{algorithmic}
\usepackage{dingbat}
\usepackage{pifont}
\usepackage{color}
\usepackage{amssymb}
\usepackage{graphicx}
\usepackage{amsmath,amssymb} 
\usepackage{amsmath,epsfig}
\usepackage{graphicx}
\usepackage{bm}
\usepackage{amssymb}
\usepackage{array}
\usepackage{extarrows}
\usepackage{url}
\usepackage{float}
\usepackage{multirow}
\usepackage{booktabs}
\usepackage{blkarray}
\usepackage{multirow}
\usepackage{array}
\usepackage{color}

\title{MAS-SAM: Segment Any Marine Animal with Aggregated Features}
\author{
Tianyu Yan$^1$
\and
Zifu Wan$^2$
\and
Xinhao Deng$^1$
\and
Pingping Zhang$^1$\thanks{Corresponding author}
\and
Yang Liu$^1$
\and
Huchuan Lu$^1$\\
\affiliations
$^1$School of Future Technology, School of Artificial Intelligence, Dalian University of Technology\\
$^2$ Robotics Institute, Carnegie Mellon University\\
\emails
\{2981431354,dengxh\}@mail.dlut.edu.cn,
zifuw@andrew.cmu.edu,
\{zhpp,ly,lhchuan\}@dlut.edu.cn
}

\begin{document}

\maketitle

\begin{abstract}
Recently, Segment Anything Model (SAM) shows exceptional performance in generating high-quality object masks and achieving zero-shot image segmentation. However, as a versatile vision model, SAM is primarily trained with large-scale natural light images. In underwater scenes, it exhibits substantial performance degradation due to the light scattering and absorption. Meanwhile, the simplicity of the SAM's decoder might lead to the loss of fine-grained object details. To address the above issues, we propose a novel feature learning framework named MAS-SAM for marine animal segmentation, which involves integrating effective adapters into the SAM's encoder and constructing a pyramidal decoder.
More specifically, we first build a new SAM's encoder with effective adapters for underwater scenes. Then, we introduce a Hypermap Extraction Module (HEM) to generate multi-scale features for a comprehensive guidance. Finally, we propose a Progressive Prediction Decoder (PPD) to aggregate the multi-scale features and predict the final segmentation results. When grafting with the Fusion Attention Module (FAM), our method enables to extract richer marine information from global contextual cues to fine-grained local details. Extensive experiments on four public MAS datasets demonstrate that our MAS-SAM can obtain better results than other typical segmentation methods.
The source code is available at https://github.com/Drchip61/MAS-SAM.
\end{abstract}
\section{Introduction}
Marine Animal Segmentation (MAS) is a critical and fundamental task in the field of visual intelligence and underwater robotics~\cite{zhang2024fantastic}. It aims at identifying and segmenting marine animals from underwater images or videos. Functionally, the accurate segmentation of marine animals is of great importance for various research areas, including marine biology, ecology and conservation. However, MAS presents unique challenges compared to typical terrestrial image segmentation~\cite{zhang2017amulet,zhang2017learning,zhang2019fast,zhang2021looking}. In fact, underwater environments are characterized by complex light scattering and absorption effects, leading to degraded image quality, reduced contrast and blurred objects. In addition, marine animals often exhibit camouflage properties, which has further complicated the segmentation task. To address these challenges, advanced perception techniques are required.

With deep Convolutional Neural Networks (CNN), many methods~\cite{zhang2018bi,zhang2019capsal,zeng2019towards,zhang2019salient,zhang2020unsupervised,liang2022weakly,deng2023recurrent} achieve significant improvements in the segmentation accuracy. However, the complex environment demands a more comprehensive understanding of underwater images to achieve accurate MAS. Meanwhile, CNN's local receptive fields are not well-suited to capture long-range dependencies and contextual information. Thus, it struggles to discern marine animals in underwater scenes.
Recently, Segment Anything Model (SAM)~\cite{kirillov2023segment} is proposed and has shown great potential in general segmentation tasks.
However, SAM's training scenarios primarily involve natural lighting conditions, which limits its performance in underwater environments. Besides, the simplistic decoder structure in SAM lacks the capability for generating fine-grained segmentation results.

Considering the above facts, in this work we propose a novel SAM-based feature learning framework named MAS-SAM for marine animal segmentation.
More specifically, by freezing the pre-trained parameters of the SAM’s encoder and introducing effective adapters, we build an Adapter-informed SAM Encoder (ASE) for extracting the unique features from marine animal images.
In addition, we construct a Hypermap Extraction Module (HEM) to extract multi-scale feature maps from the new SAM's encoder.
It serves as a comprehensive guidance for the subsequent mask prediction process.
To improve the SAM's decoder, we introduce a Progressive Prediction Decoder (PPD) to aggregate features from the original prompt, ASE and HEM.
When grafting with the Fusion Attention Module (FAM), our PPD can prioritize the importance of multi-grained feature maps and extract richer marine information from global contextual
cues to fine-grained local details.
Extensive experiments on four MAS datasets show that our method can consistently obtain outstanding results.

In summary, our contributions are as follows:
\begin{itemize}
\item We introduce a novel feature learning framework named NAS-SAM for Marine Animal Segmentation (MAS).
    It can custom SAM with aggregated multi-scale features for high-performance MAS.
\item We propose a Hypermap Extraction Module (HEM), which generates multi-scale features based on the SAM's encoder. It serves as a comprehensive guidance for the subsequent mask prediction process.
\item We propose a Progressive Prediction Decoder (PPD) to effectively improve the representation ability of SAM's decoder, capturing a wide range of marine information from global contextual cues to fine-grained local details.
\item We perform extensive experiments to verify the effectiveness of the proposed modules. Our method achieves state-of-the-art performances on four MAS datasets.
\end{itemize}
\section{Related Work}
\subsection{Marine Animal Segmentation}
MAS is a specific and challenging sub-domain of image segmentation focused on marine animals.
With the advance of deep learning, CNNs have become a popular choice for MAS.
For example, Li \emph{et al.}~\cite{li2021marine} propose an Enhanced Cascade Decoder Network (ECDNet) for accurately segmenting marine animals from complex underwater environments.
Chen \emph{et al.}~\cite{chen2022robust} propose to reduce the impact of water degradation diversity via a Siamese network.
Recently, Cheng \emph{et al.}~\cite{cheng2023bidirectional} built a bidirectional collaborative mentoring network for leveraging structural texture and contextual clues of marine images.
Fu \emph{et al.}~\cite{fu2023masnet} design a fusion network to learn semantic features of camouflage animals.
However, the lack of global understanding ability limits the performance of these CNN-based methods.
Meanwhile, vision Transformers~\cite{dosovitskiy2020image} have shown promising capabilities in capturing long-range dependencies and global contextual information.
They deliver much better performance in general image segmentation tasks~\cite{zheng2021rethinking,ranftl2021vision,liu2021tritransnet,liu2021visual}.
Inspired by these works, Hong~\emph{et al.}~\cite{hong2023usod10k} propose a hybrid network for underwater salient object segmentation, which jointly leverages the advantage of CNNs and Transformers.
However, Transformer-based methods still require massive training data to achieve satisfactory results.
Different from previous works, our method resorts to the vision fundamental model to achieve better MAS results in challenging scenes.
\subsection{SAM for Customized Tasks}
Recently, SAM~\cite{kirillov2023segment} has drawn great attention for its powerful image understanding abilities.
With appropriate prompts, it enables transfer learning across various image segmentation tasks~\cite{zhao2023enlighten,jin2023let,zhang2023segment}.
It also emerges the ability of few-shot/zero-shot segmentation, eliminating the need for task-specific retraining.
%
%
However, SAM is trained on large-scale natural images.
Thus, it fails to adequately capture the unique characteristics of other scenes and can hardly achieve outstanding performance on customized tasks.
Meanwhile, due to the decoder's simplicity, SAM only utilizes the terminal features of vision Transformers and might lead to the loss of fine-grained object details for accurate segmentation.

To relieve the above drawbacks, some works~\cite{zhang2023customized,wang2023sam,gong20233dsam,chen2023sam} have integrated domain-specific cues into SAM through the utilization of simple adapters.
Furthermore, Lai \emph{et al.}~\cite{lai2023detect} verify that sophisticated adapters can highlight task-specific information.
However, the exploration of applying SAM to MAS is rather limited.
Xu \emph{et al.}~\cite{xu2023aquasam} directly fine-tune SAM for underwater image segmentation.
Zhang \emph{et al.}~\cite{zhang2024fantastic} propose a dual-SAM structure with auto-prompts for MAS.
More importantly, previous modifications on the SAM's decoder can not capture intricate details and structures of marine animals.
Therefore, in this work we step further and custom SAM with aggregated multi-scale features for high-performance MAS.
\begin{figure*}
\centering
\resizebox{1\textwidth}{!}
{
\begin{tabular}{@{}c@{}c@{}}
\includegraphics[scale=0.9]{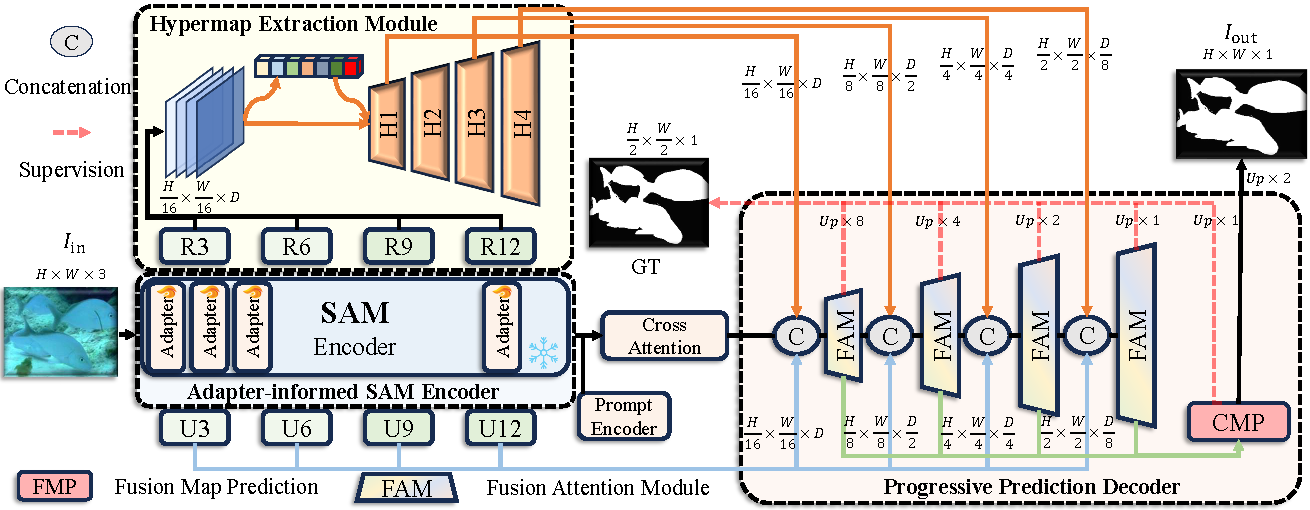} \\
\end{tabular}
}
\caption{The overall structure of our proposed framework (MAS-SAM).
It consists of three main components: Adapter-informed SAM Encoder (ASE), Hypermap Extraction Module (HEM) and Progressive Prediction Decoder (PPD).
}
\label{fig:framework}
\end{figure*}
\section{Proposed Method}
As shown in Fig.~\ref{fig:framework}, our proposed method includes three main components: Adapter-informed SAM Encoder (ASE), Hypermap Extraction Module (HEM) and Progressive Prediction Decoder (PPD).
These key components will be elaborated in the following sections.
\begin{figure}[htb]
\centering
\includegraphics[scale=0.5]{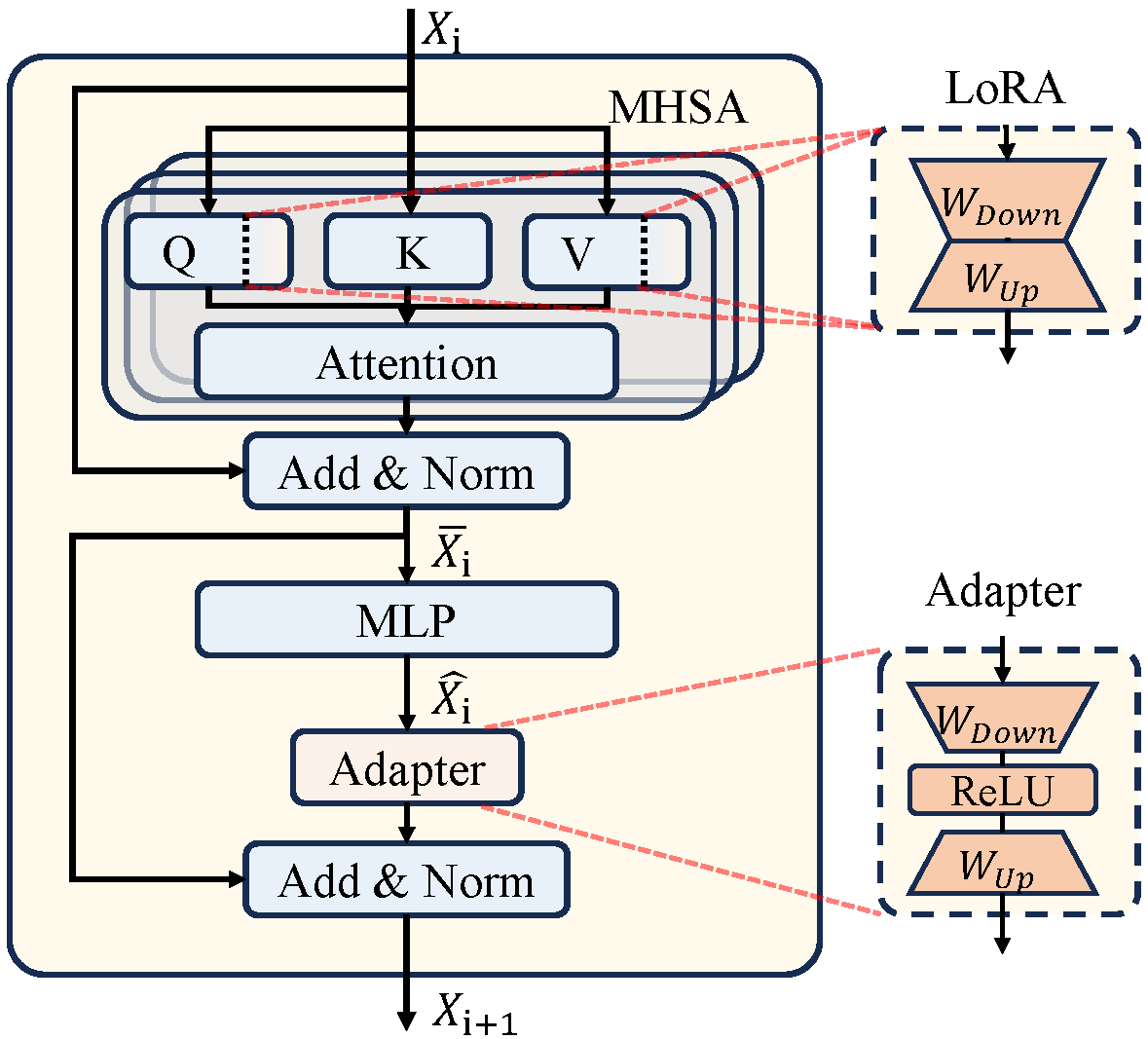}
\caption{The enhanced Transformer block in our proposed Adapter-informed SAM Encoder.}
\label{fig:adapter}
\end{figure}
\subsection{Adapter-informed SAM Encoder}
As stated in previous sections, although SAM has shown great potential for universal segmentation tasks, it is suboptimal to directly deploy the pre-trained SAM to MAS.
To address this issue, we propose an Adapter-informed SAM Encoder (ASE), which customizes SAM for marine animal images.
As shown in Fig.~\ref{fig:framework}, we retain the core components of the original SAM and utilize two parameter-efficient fine-tuning mechanisms for improving the pre-trained encoder.
As shown in Fig.~\ref{fig:adapter}, we incorporate the LoRA~\cite{hu2021lora} and Adapter~~\cite{houlsby2019parameter} into the Multi-Head Self-Attention (MHSA) and Feed-Forward
Network (FFN) of each Transformer block, respectively.
More specifically, let $\textbf{X}_{i}\in \mathbb{R}^{N\times D}$ be the input of the $i$-th Transformer block.
Here, $N$ is the number of tokens and $D$ denotes the embedding dimension.
The MHSA layer modified by LoRA can be represented as follows:
\begin{equation}\label{1}
\textbf{Q}_{i} = W_q(\textbf{X}_{i})+W^{up}_q(W^{down}_q(\textbf{X}_{i}),
\end{equation}
\vspace{-2mm}
\begin{equation}\label{3}
\textbf{K}_{i} = W_k(\textbf{X}_{i}),
\end{equation}
\vspace{-2mm}
\begin{equation}\label{2}
\textbf{V}_{i} = W_v(\textbf{X}_{i})+W^{up}_v(W^{down}_v(\textbf{X}_{i}),
\end{equation}
\vspace{-2mm}
\begin{equation}\label{4}
\overline{\textbf{X}}_{i} =MHSA\left ( \textbf{Q}_{i},\textbf{K}_{i},\textbf{V}_{i}    \right ) +  \textbf{X}_{i},
\end{equation}
where $W_{q}$, $W_{k}$ and $W_{v}$ are weights of three linear projection layers to generate the original $Query$, $Key$ and $Value$ matrices, respectively.
$W_{q,v}^{down} \in \mathbb{R}^{M\times D}$ and $W_{q,v}^{up} \in \mathbb{R}^{M\times D}$ are weights of two linear projection layers to reduce and restore the feature dimension, respectively.
$M$ is the down-projection dimension.
In this way, we can freeze the pre-trained weights ($W_{q}$, $W_{k}$ and $W_{v}$) and utilize rank decomposition matrices to greatly reduce the number of trainable parameters.

Furthermore, we insert an Adapter into the FFN as:
\begin{equation}\label{5}
\widehat{\textbf{X}}_{i}= MLP\left (LN\left ( \overline{\textbf{X}}_{i} \right )   \right ),
\end{equation}
\begin{equation}
\textbf{X}_{i+1}=W_{adpt}^{up}  \left ( \sigma \left ( W_{adpt}^{down}\left ( \widehat{\textbf{X}}_{i} \right )  \right )   \right )+\overline{\textbf{X}} _{i},
\end{equation}
where $LN$ and $MLP$ stand for the Layer Normalization (LN)~\cite{ba2016layer} and Multilayer Perceptron (MLP), respectively.
$\sigma$ is the Rectified Linear Unit (ReLU).
$W_{adpt}^{down}\in \mathbb{R}^{P\times D} $ and $W_{adpt}^{up} \in{R}^{P\times D} $ are weights of two linear projections to reduce and restore the feature dimension, respectively.
$P$ is the down-projection dimension.
Similar to LoRA, by employing an extremely low value of the parameter $P$, we can achieve a parameter-efficient fine-tuning for adapting the pre-trained SAM's encoder to marine scenes.
\subsection{Hypermap Extraction Module}
Due to the complex underwater environment, it is of vital needs to exploit both local details and global contexts for robust and accurate MAS.
As previous works point out, different Transformer layers capture different-level semantics~\cite{van2019does,bin2023unifying}.
Generally, shallow layers retain more local details and deep layers express more contextual
information.
Therefore, to enable our proposed model leverage richer marine information, we propose a Hypermap Extraction Module (HEM) to consider multi-scale feature maps from ASE.
It then serves as a comprehensive guidance for the subsequent mask prediction process.

More specifically, we first feed an image $\textbf{I}\in \mathbb{R}^{H\times W\times 3} $ into ASE and obtain the outputs of different Transformer layers.
In this work, we select the 3-6-9-12 layers and get the multi-scale token features, i.e., $\textbf{X}_i(i=3,6,9,12)$.
Then, we reshape them to spatial feature maps $\textbf{F}_i\in \mathbb{R}^{H/16\times W/16\times D}$.
To simultaneously consider these multi-scale feature maps, we perform the following feature aggregation,
\begin{equation}
\textbf{R}_{i} = \phi_{1\times1}(\textbf{F}_{i}),
\end{equation}
\begin{equation}
\overline{\textbf{H}}_{1}= \phi_{3\times3} [\textbf{R}_{3} ,\textbf{R}_{6},\textbf{R}_{9},\textbf{R}_{12} ],
\end{equation}
where $\phi_{1\times1}$ and $\phi_{3\times3}$ are convolutional layers with $1\times1$ and $3\times3$ kernels, respectively.
To improve the training stability, a Batch Normalization (BN) and a ReLU activation function are also introduced after the convolutional layers.
[\(\cdot\)] is the concatenation in channel.

Afterwards, we introduce a channel-attention layer to generate the hypermap $\textbf{H}_{j}$ as follows:
\begin{equation}\label{8}
\textbf{H}_{1}= \overline{\textbf{H}}_{1} \times \delta (\phi_{1 \times 1}(GAP(\overline{\textbf{H}}_{1})))+\overline{\textbf{H}}_{1},
\end{equation}
\begin{equation}\label{9}
\textbf{H}_{j+1}=\phi_{3\times3}(\psi_{2\times2}(\textbf{H}_{j})), j=1,2,3,
\end{equation}
where $GAP$ is the Global Average Pooling (GAP), $\delta$ is the Sigmoid function, and $\psi_{2\times2}$ is a deconvolutional layer with a $2\times2$ kernel.
As shown in the top-left part of Fig.~\ref{fig:framework}, we can obtain multi-scale hypermaps.
These hypermaps play a crucial role in improving the performance of MAS.
\subsection{Progressive Prediction Decoder}
Due to the significant appearance variations of marine animals, the simple decoder design in SAM struggles to achieve accurate segmentation masks.
To this end, inspired by \cite{zhang2020semantic,zhang2020rapnet}, we propose a Progressive Prediction Decoder (PPD) to effectively improve the prediction ability.
As shown in the right part of Fig.~\ref{fig:framework}, it has a pyramidal structure to progressively aggregate multi-source features from the original prompt, ASE and HEM, and obtain the final segmentation predictions.
\begin{figure}[h]
\centering
\includegraphics[width=1.0\linewidth, height=0.34\linewidth]{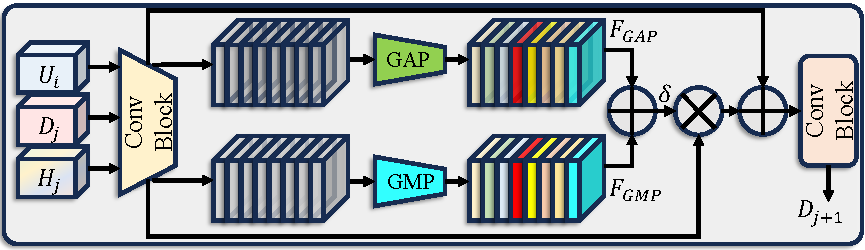}
\caption{The structure of our Fusion Attention Module.}
\label{fig:hfam}
\end{figure}

As shown in Fig.~\ref{fig:hfam}, we propose a Fusion Attention Module (FAM) to fully aggregate the multi-source features.
More specifically, we begin by up-sampling the features from ASE and resizing the input features into the same size.
Then, we fuse them as follows:
\begin{equation}
\textbf{U}_{i}=\varphi(\textbf{F}_{i}),
\end{equation}
\vspace{0.5mm}
\begin{equation}
\textbf{D}_{j+1}=FAM(\textbf{U}_{i}, \textbf{D}_{j},\textbf{H}_{j})),
\end{equation}
where $\textbf{U}_{i}$ is the up-sampled feature by utilizing a bilinear interpolation $\varphi$.
$\textbf{D}_{j}$ is the output of the $j$-th pyramidal stage in the proposed PPD.
For the FAM, we utilize a channel-attention to prioritize the importance of multi-source features.
Residual structures are also deployed to enhance the representation ability.
The procedure can be formulated as:
\begin{equation}
\textbf{Y}_{j}=\phi_{1\times1}([\textbf{U}_{i},\textbf{D}_{j},\textbf{H}_{j}]),
\end{equation}
\vspace{0.5mm}
\begin{equation}\label{6}
\textbf{W}_{j}=\delta(\phi_{1 \times 1}(GAP(\textbf{Y}_{j}))+\phi_{1 \times 1}(GMP(\textbf{Y}_{j}))),
\end{equation}
\vspace{0.5mm}
\begin{equation}\label{7}
\overline{\textbf{D}}_{j+1}=W_{j}(\textbf{Y}_{j})+\textbf{Y}_{j},
\end{equation}
\begin{equation}\label{8}
\textbf{D}_{j+1}=\phi_{1\times1}(\overline{\textbf{D}}_{j+1}),
\end{equation}
where $GMP$ is the Global Max Pooling (GMP).
The channel-wise weights can highlight relevant features and suppress irrelevant ones~\cite{yan2023transy}.
Meanwhile, the attention mechanism employed by our FAM helps to capture intricate relationships between features across various scales, resulting in more coherent and informative feature representations.
Thus, FAM can effectively integrate and refine the multi-source features.

Finally, to achieve the progressive prediction, we build the PPD grafted with the FAM as follows:
\begin{equation}
\textbf{P}_{j}= \phi_{1\times1}(\varphi(FAM(\textbf{U}_{i}, \textbf{D}_{j},\textbf{H}_{j}))),
\end{equation}
where $\textbf{P}_{j}$ is the prediction mask of the $j$-th pyramidal stage.
The proposed PPD facilitates the seamless aggregation of multi-source features from the original
prompt, ASE and HEM, resulting in richer marine information from global contextual cues to fine-grained local details.

To further improve the prediction results, we take all the predictions at different stages and generate the final prediction as follows:
\begin{equation}
\textbf{P}= \phi_{1\times1}([\textbf{P}_{1},\textbf{P}_{2},\textbf{P}_{3},\textbf{P}_{4}]).
\end{equation}
With the synergistic use of the pyramid structure and FAM, our MAS-SAM can effectively leverage the diverse information and produce highly refined and detailed segmentation masks for a wide range of marine animal shapes and sizes.
\subsection{Model Training}
During training, we follow previous methods~\cite{li2021marine,yan2022fully} and use deep supervision from three levels, i.e., pixel-level supervision (Binary Cross-Entropy Loss), region-level supervision (SSIM Loss) and overall-level supervision (IoU Loss).
Thus, we define $\mathcal{L}^{f}$ or $\mathcal{L}^{j}$ as a combined loss with three terms:
\begin{equation}\label{13}
\mathcal{L}^{f/j} = \mathcal{L}_{BCE}+\mathcal{L}_{SSIM}+\mathcal{L}_{IoU},
\end{equation}
where $\mathcal{L}^{f}$ and $\mathcal{L}^{j}$ are losses of the final prediction and the $j$-th stage-output, respectively.
Due to the space limitation, we refer readers to~\cite{yan2022fully} for the exact formulas.
\begin{table*}[h]
\renewcommand\arraystretch{1.1}
\setlength\tabcolsep{5.5pt}
\centering
\resizebox{0.90\textwidth}{!}
{
\begin{tabular}{c|c|c|c|c|c|c|c|c|c|c}
\hline
&\multicolumn{5}{c|}{\textbf{MAS3K}}        &\multicolumn{5}{c}{\textbf{RMAS}} \\ \cline{2-11}
\textbf{Method}&\textbf{mIoU} & $\textbf{S}_\alpha$ & $\textbf{F}_\beta^w $&$ \textbf{mE}_\phi$ & \textbf{MAE}& \textbf{mIoU} & $\textbf{S}_\alpha$ & $\textbf{F}_\beta^w$ & $\textbf{mE}_\phi$ & \textbf{MAE} \\
\hline
 UNet++~\cite{zhou2018unet++}   &0.506&0.726&0.552&0.790&0.083&0.558&0.763&0.644&0.835&0.046\\
 BASNet~\cite{qin2019basnet}   &0.677&0.826&0.724&0.862&0.046&0.707&0.847&0.771&0.907&0.032\\
 PFANet~\cite{zhao2019pyramid}   &0.405&0.690&0.471&0.768&0.086&0.556&0.767&0.582&0.810&0.051\\
 SCRN~\cite{wu2019stacked}   &0.693&0.839&0.730&0.869&0.041&0.695&0.842&0.731&0.878&0.030\\
 U2Net~\cite{qin2020u2}   &0.654&0.812&0.711&0.851&0.047&0.676&0.830&0.762&0.904&0.029\\
 SINet~\cite{fan2020camouflaged} &0.658&0.820&0.725&0.884&0.039&0.684&0.835&0.780&0.908&0.025\\
 PFNet~\cite{mei2021camouflaged}  &0.695&0.839&0.746&0.890&0.039&0.694&0.843&0.771&0.922&0.026\\
 RankNet~\cite{lv2021simultaneously} &0.658&0.812&0.722&0.867&0.043&0.704&0.846&0.772&0.927&0.026\\
 C2FNet~\cite{sun2021context}   &0.717&0.851&0.761&0.894&0.038&0.721&0.858&0.788&0.923&0.026\\
 ECDNet~\cite{li2021marine}   &0.711&0.850&0.766&0.901&0.036&0.664&0.823&0.689&0.854&0.036\\
 OCENet~\cite{liu2022modeling}  &0.667&0.824&0.703&0.868&0.052&0.680&0.836&0.752&0.900&0.030\\
 ZoomNet~\cite{pang2022zoom}   &0.736&0.862&0.780&0.898&0.032&0.728&0.855&0.795&0.915&0.022\\
 MASNet~\cite{fu2023masnet} &0.742&0.864&0.788&0.906&0.032&\underline{0.731}&\underline{0.862}&\underline{0.801}&0.920&0.024\\
 \hline
 SETR~\cite{zheng2021rethinking}   &0.715&0.855&0.789&0.917&0.030&0.654&0.818&0.747&0.933&0.028\\
 TransUNet~\cite{chen2021transunet}   &0.739&0.861&0.805&0.919&0.029&0.688&0.832&0.776&\underline{0.941}&0.025\\
 H2Former~\cite{he2023h2former}   &\underline{0.748}&0.865&0.810&\underline{0.925}&\underline{0.028}&0.717&0.844&0.799&0.931&\underline{0.023}\\
 \hline
 SAM~\cite{kirillov2023segment}  &0.566&0.763&0.656&0.807&0.059&0.445&0.697&0.534&0.790&0.053\\
 I-MedSAM\cite{wei2023medsam}   &0.698&0.835&0.759&0.889&0.039&0.633&0.803&0.699&0.893&0.035\\
 Med-SAM\cite{wu2023medical}   &0.739& 0.861& \underline{0.811} &0.922& 0.031&0.678&0.832&0.778&0.920&0.027\\
 SAM-Adapter\cite{chen2023sam}   &0.714&0.847&0.782&0.914&0.033&0.656&0.816&0.752&0.927&0.027\\
 SAM-DADF~\cite{lai2023detect}   &0.742&\underline{0.866}&0.806&0.925&0.028&0.686&0.833&0.780&0.926&0.024\\
 \hline
\textbf{MAS-SAM}  &\textbf{0.788}&\textbf{0.887}&\textbf{0.840}&\textbf{0.938}&\textbf{0.025}&\textbf{0.742}&\textbf{0.865}&\textbf{0.819}&\textbf{0.948}&\textbf{0.021}\\
\hline
\end{tabular}
}
\caption{Performance comparison on MAS3K and RMAS. The best and second results are in bold and underlined, respectively.}
\label{mas3k}
\end{table*}
\section{Experiments}
\subsection{Datasets and Evaluation Metrics}
In this work, we adopt four public MAS benchmarks to evaluate the model performance.
The \textbf{MAS3K} dataset~\cite{li2020mas3k} comprises 3,103 marine images, in which 193 are background images. Following the default split, we use 1,769 images for training and 1,141 images for testing.
The \textbf{RMAS} dataset~\cite{fu2023masnet} consists of 3,014 marine animal images with 2,514 images used for training and 500 images for testing.
The \textbf{UFO120} dataset~\cite{islam2020simultaneous} comprises 1,620 underwater images with various scenes. Following the default split, we use 1,500 images for training and 120 images for testing.
The \textbf{RUWI} dataset~\cite{drews2021underwater} is a real underwater image dataset captured under complex light conditions. The dataset consists of 700 images, splitting into 525 images for training and 175 images for testing.

Meanwhile, we adopt five metrics to evaluate the segmentation performance of different models.
They are Mean Intersection over Union (mIoU), Structural Similarity Measure ($S_\alpha$)~\cite{fan2017structure}, Weighted F-measure ($F_\beta^w$)~\cite{margolin2014evaluate}, Mean Enhanced-Alignment Measure ($mE_\phi$)~\cite{fan2021cognitive} and mean absolute error (MAE).
\subsection{Implementation Details}
Our model is implemented with the PyTorch toolbox and one RTX 3090 GPU.
For our model, the SAM's encoder and prompt encoder are initialized from the pre-trained SAM-B~\cite{kirillov2023segment}, while other proposed modules are randomly initialized.
During the training process, we freeze the SAM's encoder and only fine-tune the remaining modules.
The widely-used AdamW optimizer~\cite{loshchilov2017decoupled} is adopted to update the model parameters.
The initial learning rate and weight decay are set to 0.001 and 0.1, respectively.
We reduce the learning rate by a factor of 10 at every 20 epochs.
The total number of training epochs is set to 50.
The mini-batch size is set to 8.
The input images are uniformly resized to $512\times512\times3$.
%
\begin{table*}[h]
\renewcommand\arraystretch{1.1}
\setlength\tabcolsep{5.5pt}
\centering
\resizebox{0.90\textwidth}{!}
{
\begin{tabular}{c|c|c|c|c|c|c|c|c|c|c}
\hline
&\multicolumn{5}{c|}{\textbf{UFO120}}        &\multicolumn{5}{c}{\textbf{RUWI}} \\ \cline{2-11}
\textbf{Method}&\textbf{mIoU} & $\textbf{S}_\alpha$ & $\textbf{F}_\beta^w$ & $\textbf{m}\textbf{E}_\phi$ & \textbf{MAE}& \textbf{mIoU} & $\textbf{S}_\alpha$ & $\textbf{F}_\beta^w$ & $\textbf{m}\textbf{E}_\phi$ & \textbf{MAE} \\
\hline
    UNet++~\cite{zhou2018unet++}    &0.412&0.459&0.433&0.451&0.409&0.586&0.714&0.678&0.790&0.145\\
    BASNet~\cite{qin2019basnet}   &0.710&0.809&0.793&0.865&0.097&0.841&0.871&0.895&0.922&0.056\\
    PFANet~\cite{zhao2019pyramid}   &0.677&0.752&0.723&0.815&0.129&0.773&0.765&0.811&0.867&0.096\\
    SCRN~\cite{wu2019stacked}   &0.678&0.783&0.760&0.839&0.106&0.830&0.847&0.883&0.925&0.059\\
    U2Net~\cite{qin2020u2}   &0.680&0.792&0.709&0.811&0.134&0.841&0.873&0.861&0.786&0.074\\
    SINet~\cite{fan2020camouflaged}   &0.767&0.837&0.834&0.890&0.079&0.785&0.789&0.825&0.872&0.096\\
    PFNet~\cite{mei2021camouflaged}   &0.570&0.708&0.550&0.683&0.216&0.864&0.883&0.870&0.790&0.062\\
    RankNet~\cite{lv2021simultaneously}   &0.739&0.823&0.772&0.828&0.101&0.865&0.886&0.889&0.759&0.056\\
    C2FNet~\cite{sun2021context}   &0.747&0.826&0.806&0.878&0.083&0.840&0.830&0.883&0.924&0.060\\
    ECDNet~\cite{li2021marine}   &0.693&0.783&0.768&0.848&0.103&0.829&0.812&0.871&0.917&0.064\\
    OCENet~\cite{liu2022modeling}  &0.605&0.725&0.668&0.773&0.161&0.763&0.791&0.798&0.863&0.115\\
    ZoomNet~\cite{pang2022zoom}   &0.616&0.702&0.670&0.815&0.174&0.739&0.753&0.771&0.817&0.137\\
    MASNet~\cite{fu2023masnet} &0.754&0.827&0.820&0.879&0.083&0.865&0.880&0.913&0.944&0.047\\
    \hline

    SETR~\cite{zheng2021rethinking}   &0.711&0.811&0.796&0.871&0.089&0.832&0.864&0.895&0.924&0.055\\
    TransUNet~\cite{chen2021transunet}   &0.752&0.825&0.827&0.888&0.079&0.854&0.872&0.910&0.940&0.048\\

    H2Former~\cite{he2023h2former}   &\underline{0.780}&\underline{0.844}&\underline{0.845}&\underline{0.901}&\underline{0.070}&0.871&0.884&0.919&0.945&0.045\\
    \hline
    SAM~\cite{kirillov2023segment}   &0.681&0.768&0.745&0.827&0.121&0.849&0.855&0.907&0.929&0.057\\
    I-MedSAM~\cite{wei2023medsam}   &0.730&0.818&0.788&0.865&0.084&0.844&0.849&0.897&0.923&0.050\\
    Med-SAM~\cite{wu2023medical}   &0.774& 0.842& 0.839 &0.899& 0.072&0.877&0.885&0.921&0.942&0.045\\
    SAM-Adapter~\cite{chen2023sam}   &0.757&0.829&0.834&0.884&0.081&0.867&0.878&0.913&\underline{0.946}&0.046\\
    SAM-DADF~\cite{lai2023detect}   &0.768&0.841&0.836&0.893&0.073&\underline{0.881}&\underline{0.889}&\underline{0.925}&0.940&\underline{0.044}\\
    \hline
\textbf{MAS-SAM}   &\textbf{0.807}&\textbf{0.861}&\textbf{0.864}&\textbf{0.914}&\textbf{0.063}&\textbf{0.902}&\textbf{0.894}&\textbf{0.941}&\textbf{0.961}&\textbf{0.035}\\
\hline
\end{tabular}
}
\caption{Performance comparison on UFO120 and RUWI. The best and second results are in bold and underlined, respectively.}
\label{ufo}
\end{table*}
\subsection{Comparison with State-of-the-arts}
In this section, we compare our method with other approaches on four public MAS datasets.
Specifically, we follow previous works,~\emph{i.e.}, MASNet~\cite{fu2023masnet} and ECDNet~\cite{li2021marine}, to include compared methods/tools.
MASNet and ECDNet are typical MAS methods.
Other compared methods are modified for underwater scene segmentation.
Especially, some methods (\emph{e.g.}, SINet~\cite{fan2020camouflaged}, C2FNet~\cite{sun2021context}) focus on camouflage object detection for underwater images.
Thus, it is very reasonable to compare with them.
I-MedSAM~\cite{wei2023medsam}, MedSAM~\cite{wu2023medical}, SAM-Adapter~\cite{chen2023sam} and SAM-DADF~\cite{lai2023detect} are similar to our method in addressing the adaptability issue of SAM for downstream tasks.

\textbf{Quantitative Comparisons.}
Tab.~\ref{mas3k} and Tab.~\ref{ufo} illustrate the quantitative results of compared methods.
They clearly demonstrate that our method outperforms other methods on all five metrics across four datasets.
These results show the superiority of our method in terms of overall completeness, structural accuracy, and pixel-wise precision.

Compared with CNN-based methods, our method delivers significant advantages in multi-scale and long-range modeling capabilities.
On the large-scale MAS3K dataset, our method achieves the best mIoU, $S_\alpha$ , $F_\beta^w$ and $mE_\phi$ values.
It demonstrates an improvement of approximately 3-4\% across various metrics.
Compared with Transformer-based methods, our method shows a 2-3\% gain across multiple metrics on the MAS3K dataset.
Moreover, significant improvements are also observed on other three datasets.
Compared with SAM-based methods, our method efficiently injects specific marine scene information into SAM through two concise adapter modules.
In short, our design notably enhances the model's segmentation ability and achieves a considerable performance gain over other approaches.

\begin{figure*}[!t]
\centering
\resizebox{0.94\textwidth}{!}
{
\renewcommand\arraystretch{0.1}
\begin{tabular}{@{}c@{}c@{}c@{}c@{}c@{}c@{}c@{}c@{}c@{}c@{}c@{}c@{}c@{}c@{}c@{}c}
\vspace{0.5mm}
\includegraphics[width=0.0909\linewidth,height=1.2cm]{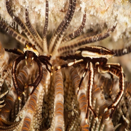}\ &
\includegraphics[width=0.0909\linewidth,height=1.2cm]{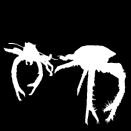}\ &
\includegraphics[width=0.0909\linewidth,height=1.2cm]{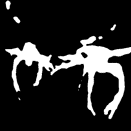}\ &
\includegraphics[width=0.0909\linewidth,height=1.2cm]{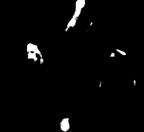}\ &
\includegraphics[width=0.0909\linewidth,height=1.2cm]{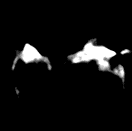}\ &
\includegraphics[width=0.0909\linewidth,height=1.2cm]{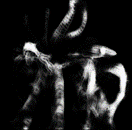}\ &
\includegraphics[width=0.0909\linewidth,height=1.2cm]{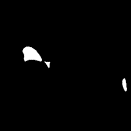}\ &
\includegraphics[width=0.0909\linewidth,height=1.2cm]{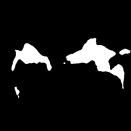}\ &
\includegraphics[width=0.0909\linewidth,height=1.2cm]{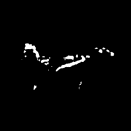}\ &
\includegraphics[width=0.0909\linewidth,height=1.2cm]{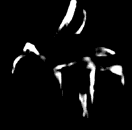}\ &
\includegraphics[width=0.0909\linewidth,height=1.2cm]{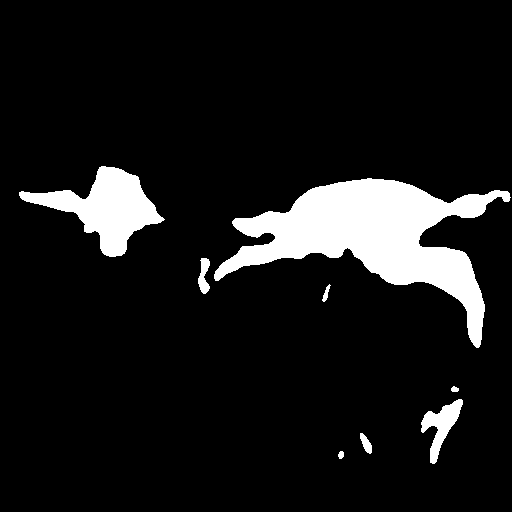}\  \\
\vspace{0.5mm}
\includegraphics[width=0.0909\linewidth,height=1.2cm]{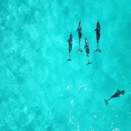}\ &
\includegraphics[width=0.0909\linewidth,height=1.2cm]{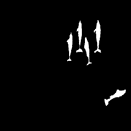}\ &
\includegraphics[width=0.0909\linewidth,height=1.2cm]{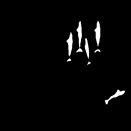}\ &
\includegraphics[width=0.0909\linewidth,height=1.2cm]{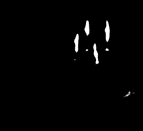}\ &
\includegraphics[width=0.0909\linewidth,height=1.2cm]{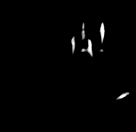}\ &
\includegraphics[width=0.0909\linewidth,height=1.2cm]{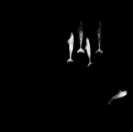}\ &
\includegraphics[width=0.0909\linewidth,height=1.2cm]{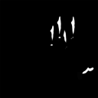}\ &
\includegraphics[width=0.0909\linewidth,height=1.2cm]{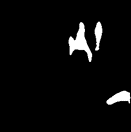}\ &
\includegraphics[width=0.0909\linewidth,height=1.2cm]{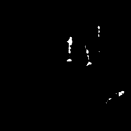}\ &
\includegraphics[width=0.0909\linewidth,height=1.2cm]{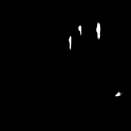}\ &
\includegraphics[width=0.0909\linewidth,height=1.2cm]{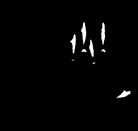}\  \\
\vspace{0.5mm}
\includegraphics[width=0.0909\linewidth,height=1.2cm]{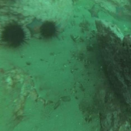}\ &
\includegraphics[width=0.0909\linewidth,height=1.2cm]{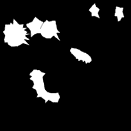}\ &
\includegraphics[width=0.0909\linewidth,height=1.2cm]{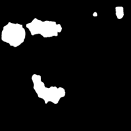}\ &
\includegraphics[width=0.0909\linewidth,height=1.2cm]{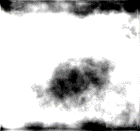}\ &
\includegraphics[width=0.0909\linewidth,height=1.2cm]{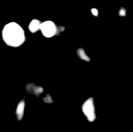}\ &
\includegraphics[width=0.0909\linewidth,height=1.2cm]{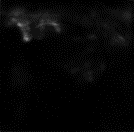}\ &
\includegraphics[width=0.0909\linewidth,height=1.2cm]{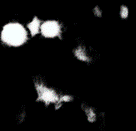}\ &
\includegraphics[width=0.0909\linewidth,height=1.2cm]{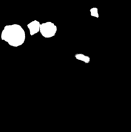}\ &
\includegraphics[width=0.0909\linewidth,height=1.2cm]{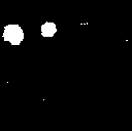}\ &
\includegraphics[width=0.0909\linewidth,height=1.2cm]{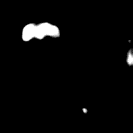}\ &
\includegraphics[width=0.0909\linewidth,height=1.2cm]{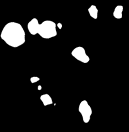}\  \\
\vspace{0.5mm}
\includegraphics[width=0.0909\linewidth,height=1.2cm]{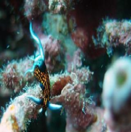}\ &
\includegraphics[width=0.0909\linewidth,height=1.2cm]{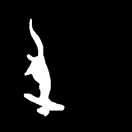}\ &
\includegraphics[width=0.0909\linewidth,height=1.2cm]{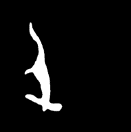}\ &
\includegraphics[width=0.0909\linewidth,height=1.2cm]{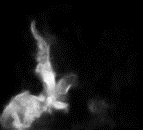}\ &
\includegraphics[width=0.0909\linewidth,height=1.2cm]{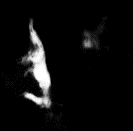}\ &
\includegraphics[width=0.0909\linewidth,height=1.2cm]{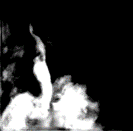}\ &
\includegraphics[width=0.0909\linewidth,height=1.2cm]{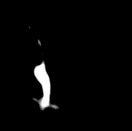}\ &
\includegraphics[width=0.0909\linewidth,height=1.2cm]{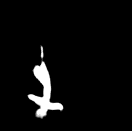}\ &
\includegraphics[width=0.0909\linewidth,height=1.2cm]{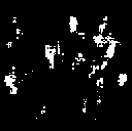}\ &
\includegraphics[width=0.0909\linewidth,height=1.2cm]{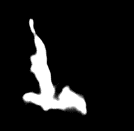}\ &
\includegraphics[width=0.0909\linewidth,height=1.2cm]{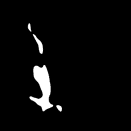}\  \\
\vspace{0.5mm}
\includegraphics[width=0.0909\linewidth,height=1.2cm]{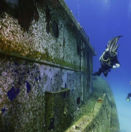}\ &
\includegraphics[width=0.0909\linewidth,height=1.2cm]{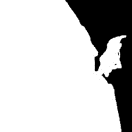}\ &
\includegraphics[width=0.0909\linewidth,height=1.2cm]{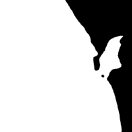}\ &
\includegraphics[width=0.0909\linewidth,height=1.2cm]{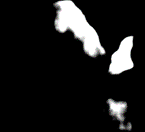}\ &
\includegraphics[width=0.0909\linewidth,height=1.2cm]{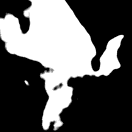}\ &
\includegraphics[width=0.0909\linewidth,height=1.2cm]{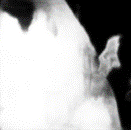}\ &
\includegraphics[width=0.0909\linewidth,height=1.2cm]{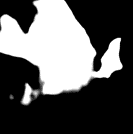}\ &
\includegraphics[width=0.0909\linewidth,height=1.2cm]{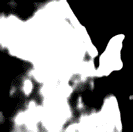}\ &
\includegraphics[width=0.0909\linewidth,height=1.2cm]{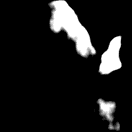}\ &
\includegraphics[width=0.0909\linewidth,height=1.2cm]{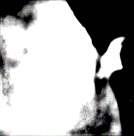}\ &
\includegraphics[width=0.0909\linewidth,height=1.2cm]{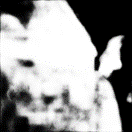}\  \\
\vspace{0.5mm}
\includegraphics[width=0.0909\linewidth,height=1.2cm]{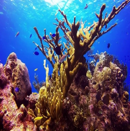}\ &
\includegraphics[width=0.0909\linewidth,height=1.2cm]{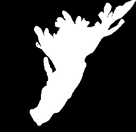}\ &
\includegraphics[width=0.0909\linewidth,height=1.2cm]{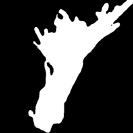}\ &
\includegraphics[width=0.0909\linewidth,height=1.2cm]{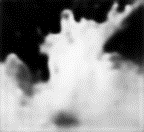}\ &
\includegraphics[width=0.0909\linewidth,height=1.2cm]{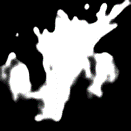}\ &
\includegraphics[width=0.0909\linewidth,height=1.2cm]{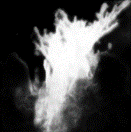}\ &
\includegraphics[width=0.0909\linewidth,height=1.2cm]{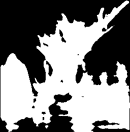}\ &
\includegraphics[width=0.0909\linewidth,height=1.2cm]{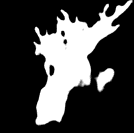}\ &
\includegraphics[width=0.0909\linewidth,height=1.2cm]{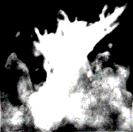}\ &
\includegraphics[width=0.0909\linewidth,height=1.2cm]{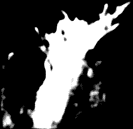}\ &
\includegraphics[width=0.0909\linewidth,height=1.2cm]{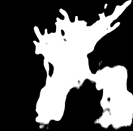}\  \\
\vspace{0.5mm}
\includegraphics[width=0.0909\linewidth,height=1.2cm]{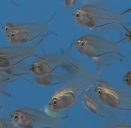}\ &
\includegraphics[width=0.0909\linewidth,height=1.2cm]{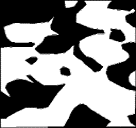}\ &
\includegraphics[width=0.0909\linewidth,height=1.2cm]{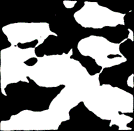}\ &
\includegraphics[width=0.0909\linewidth,height=1.2cm]{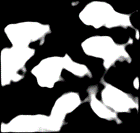}\ &
\includegraphics[width=0.0909\linewidth,height=1.2cm]{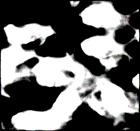}\ &
\includegraphics[width=0.0909\linewidth,height=1.2cm]{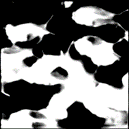}\ &
\includegraphics[width=0.0909\linewidth,height=1.2cm]{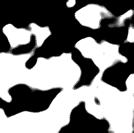}\ &
\includegraphics[width=0.0909\linewidth,height=1.2cm]{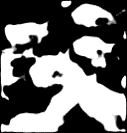}\ &
\includegraphics[width=0.0909\linewidth,height=1.2cm]{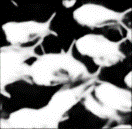}\ &
\includegraphics[width=0.0909\linewidth,height=1.2cm]{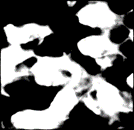}\ &
\includegraphics[width=0.0909\linewidth,height=1.2cm]{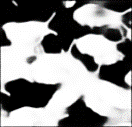}\  \\
\vspace{0.5mm}
\includegraphics[width=0.0909\linewidth,height=1.2cm]{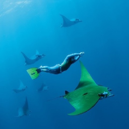}\ &
\includegraphics[width=0.0909\linewidth,height=1.2cm]{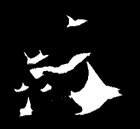}\ &
\includegraphics[width=0.0909\linewidth,height=1.2cm]{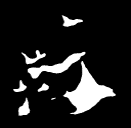}\ &
\includegraphics[width=0.0909\linewidth,height=1.2cm]{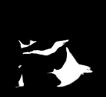}\ &
\includegraphics[width=0.0909\linewidth,height=1.2cm]{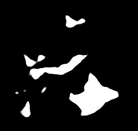}\ &
\includegraphics[width=0.0909\linewidth,height=1.2cm]{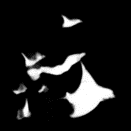}\ &
\includegraphics[width=0.0909\linewidth,height=1.2cm]{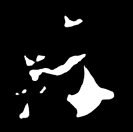}\ &
\includegraphics[width=0.0909\linewidth,height=1.2cm]{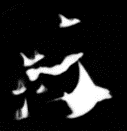}\ &
\includegraphics[width=0.0909\linewidth,height=1.2cm]{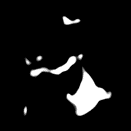}\ &
\includegraphics[width=0.0909\linewidth,height=1.2cm]{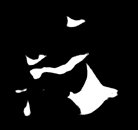}\ &
\includegraphics[width=0.0909\linewidth,height=1.2cm]{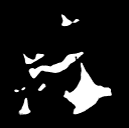}\  \\
\vspace{0.5mm}
 {\small Image} & {\small GT} & {\small MAS-SAM} & {\small ECDNet}& {\small MASNet}  & {\small SETR} & {\small TransUNet} & {\small H2Former} & {\small SAM} & {\small SAM-Adapt}& {\small SAM-DADF} \\
\end{tabular}
}
\caption{Visual comparison of predicted segmentation masks with different methods. The images in the 1-2nd rows are from the MAS3K, 3-4th are from RMAS, 5-6th are from UFO120 and the 7-8th are from RUWI. Best view by zooming in.}
\label{fig:visual}
\end{figure*}

\textbf{Qualitative Comparisons.}
To demonstrate the superiority of our method more intuitively, we present visual comparisons of different methods in Fig.~\ref{fig:visual}.
The visualizations provide a clearer advantage of our method compared with the previous approaches.
There are challenging images with high background clutters and abundant details.
However, our method still generates better segmentation masks.
\subsection{Ablation Studies}
On the MAS3K dataset, we conduct experiments to verify the effectiveness of each module in our model.
More results are in the supplementary material.
The model (A) is directly deploying the pre-trained SAM.
Visual results are in Fig.~\ref{fig:ablation}.
\begin{table*}[h]
    \centering
    \renewcommand\arraystretch{1.1}
    \setlength\tabcolsep{5.5pt}
    \resizebox{0.8\textwidth}{!}
{
    \begin{tabular}{c|cccccc|c|c|c|c|c}
        \hline
        &\multicolumn{6}{c|}{\textbf{Method}}        &\multicolumn{5}{c}{\textbf{MAS3K}} \\ \cline{2-12}
        & \textbf{LoRA}& \textbf{Adapter}& \textbf{FPN}& \textbf{HEM}& \textbf{FAM}& \textbf{ML} & \textbf{mIoU} & $\textbf{S}_\alpha$ & $\textbf{F}_\beta^w$ & $\textbf{m}\textbf{E}_\phi$ & \textbf{MAE} \\
        \hline
        (A) & \ding{53}& \ding{53}& \ding{53}& \ding{53}& \ding{53}& \ding{53}& 0.566 & 0.763 & 0.656 & 0.807& 0.059 \\
        (B) & \ding{51}& \ding{53}& \ding{53}& \ding{53}& \ding{53}& \ding{53}& 0.742 & 0.867 & 0.806& 0.919& 0.029 \\
        (C) & \ding{51}& \ding{51}& \ding{53}& \ding{53}& \ding{53}& \ding{53}& 0.755 & 0.868 & 0.813 & 0.924& 0.028 \\
        (D) & \ding{51}& \ding{51}& \ding{51}& \ding{53}& \ding{53}& \ding{53}& 0.763 & 0.871 & 0.819 & 0.922& 0.028\\
        (E) & \ding{51}& \ding{51}& \ding{51}& \ding{51}& \ding{53}& \ding{53}& 0.771&0.875&	0.827&	0.926& 0.028 \\
        (F) & \ding{51}& \ding{51}& \ding{51}& \ding{51}& \ding{51}& \ding{53}& 0.781 & 0.881 & 0.834 & 0.933& 0.025 \\
        (G) & \ding{51}& \ding{51}& \ding{51}& \ding{51}& \ding{51}& \ding{51}& 0.788 & 0.887 & 0.840 & 0.938& 0.025 \\
        \hline
    \end{tabular}
    }
    \caption{Performance comparisons of using different modules on MAS3K dataset.}
    \label{ablation}
\end{table*}
\begin{figure*}[!t]
\centering
\resizebox{0.96\textwidth}{!}
{
\renewcommand\arraystretch{0.1}
\begin{tabular}{@{}c@{}c@{}c@{}c@{}c@{}c@{}c@{}c@{}c@{}c@{}c@{}c@{}c@{}c}
\vspace{0.5mm}
\includegraphics[width=0.11\linewidth,height=1.2cm]{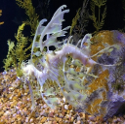}\ &
\includegraphics[width=0.11\linewidth,height=1.2cm]{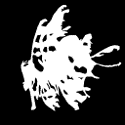}\ &
\includegraphics[width=0.11\linewidth,height=1.2cm]{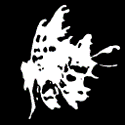}\ &
\includegraphics[width=0.11\linewidth,height=1.2cm]{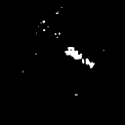}\ &
\includegraphics[width=0.11\linewidth,height=1.2cm]{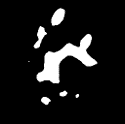}\ &
\includegraphics[width=0.11\linewidth,height=1.2cm]{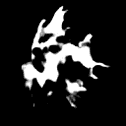}\ &
\includegraphics[width=0.11\linewidth,height=1.2cm]{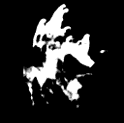}\ &
\includegraphics[width=0.11\linewidth,height=1.2cm]{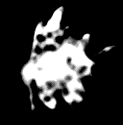}\ &
\includegraphics[width=0.11\linewidth,height=1.2cm]{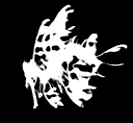}\  \\
\vspace{0.5mm}
\includegraphics[width=0.11\linewidth,height=1.2cm]{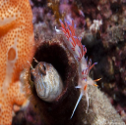}\ &
\includegraphics[width=0.11\linewidth,height=1.2cm]{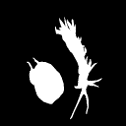}\ &
\includegraphics[width=0.11\linewidth,height=1.2cm]{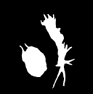}\ &
\includegraphics[width=0.11\linewidth,height=1.2cm]{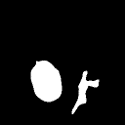}\ &
\includegraphics[width=0.11\linewidth,height=1.2cm]{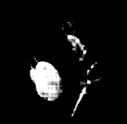}\ &
\includegraphics[width=0.11\linewidth,height=1.2cm]{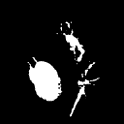}\ &
\includegraphics[width=0.11\linewidth,height=1.2cm]{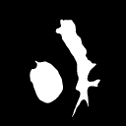}\ &
\includegraphics[width=0.11\linewidth,height=1.2cm]{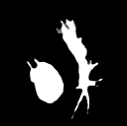}\ &
\includegraphics[width=0.11\linewidth,height=1.2cm]{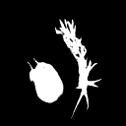}\  \\
\vspace{0.5mm}
\includegraphics[width=0.11\linewidth,height=1.2cm]{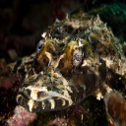}\ &
\includegraphics[width=0.11\linewidth,height=1.2cm]{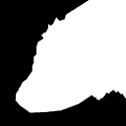}\ &
\includegraphics[width=0.11\linewidth,height=1.2cm]{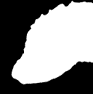}\ &
\includegraphics[width=0.11\linewidth,height=1.2cm]{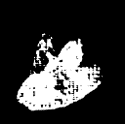}\ &
\includegraphics[width=0.11\linewidth,height=1.2cm]{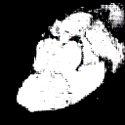}\ &
\includegraphics[width=0.11\linewidth,height=1.2cm]{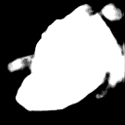}\ &
\includegraphics[width=0.11\linewidth,height=1.2cm]{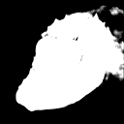}\ &
\includegraphics[width=0.11\linewidth,height=1.2cm]{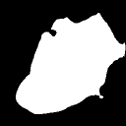}\ &
\includegraphics[width=0.11\linewidth,height=1.2cm]{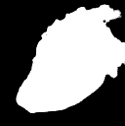}\  \\
\vspace{0.5mm}
 {\small Image} & {\small GT} & {\small MAS-SAM} & {\small SAM}  & {\small LoRA} & {\small ASE} & {\small FPN} & {\small HEM} & {\small PPD}  \\
\end{tabular}
}
\caption{Visual comparison of predicted segmentation masks with different modules. Best view by zooming in.}
\label{fig:ablation}
\end{figure*}

\textbf{Effects of Adapters in ASE.}
In the 1-3 rows of Tab.~\ref{ablation}, we show the impact of adopters in the ASE. By incorporating two effective adapters, our method has a significant improvement in terms of all evaluation metrics.
These results indicate that it is very necessary and powerful to inject marine environment information into the fundamental model.

\textbf{Effects of Key Modules.}
In the 3-6 rows of Tab.~\ref{ablation}, we examine the impact of other key modules.
By constructing a PPD, the model can capture more local details, resulting in a 0.8\% increase in mIoU.
Subsequently, HEM enables the full utilization of multi-level features extracted by the ASE, serving as informative guidance for the subsequent prediction structure. This enhancement leads to a noticeable improvement in performance, better than typical FPN structures. Lastly, we leverage the fused information through the FAM. Compared with the original SAM decoder, our method achieves great improvements in all metrics.

\textbf{Effects of Different Losses.}
In the 6-7 rows of Tab.~\ref{ablation}, we demonstrate the significance of utilizing multi-level supervision during the training process. By employing three levels of supervision, our method shows a notable improvement.
\subsection{More Discussions}
In this work, our aim is to adapt SAM to MAS.
However, there are three key issues. 1) Domain differences. SAM is pre-trained on natural light scenes, while
MAS always suffers from the refraction and absorption of light.
Thus, directly employing SAM is hard to achieve good performance for MAS.
2) Camouflage behaviours. Marine animals usually exhibit camouflage behaviours, and large variations in shapes and sizes.
The original SAM is not good at modeling these appearances.
3) Missing details. With cross-attentions and only two deconvolutions, the original SAM’s decoder leads to a substantial loss of object details.

To address above issues, we present three contributions as follows:
1) We inject marine domain knowledge into the SAM backbone through efficient adapters, thus making SAM more adaptable to marine scene tasks.
2) We employ a HEM to fully leverage multi-scale information.
This helps our model to accurately locate marine animals against complex appearances.
3) We propose a PPD structure to progressively integrate feature maps from different levels.
With three-grained supervision losses, it can capture the intricate details of marine life extensively.
The above facts make our work valuable for underwater intelligence.
%
\section{Conclusion}
In this work, we propose a novel feature learning framework named MAS-SAM for MAS.
Specifically, we first propose an Adapter-informed SAM Encoder (ASE) for underwater scenes.
Then, we introduce a Hypermap Extraction Module (HEM) to generate multi-scale features for a
comprehensive guidance.
Finally, we propose a Progressive Prediction Decoder (PPD) to aggregate the multi-scale features and predict the final segmentation results.
When grafting with the Fusion Attention Module (FAM), our approach extracts richer marine information from global contextual cues to fine-grained local details.
Extensive experiments on four MAS datasets show the effectiveness of our MAS-SAM.
\section*{Acknowledgements}
This work was supported in part by the National Natural Science Foundation of China (No. 62101092), the Fundamental Research Funds for the Central Universities (No. DUT23YG232) and the Open Project Program of State Key Laboratory of Virtual Reality Technology and Systems, Beihang University (No. VRLAB2022C02).

\small
\bibliographystyle{named}
\bibliography{ijcai24}
\end{document}